\documentclass[journal, twocolumn, 11 pt]{IEEEtran}
\newlength{\tocsep}
\usepackage[utf8]{inputenc}
\usepackage{verbatim}
\usepackage{float}
\usepackage{url}
\usepackage{array}
\usepackage{color}
\usepackage{multirow,array,graphicx}
\usepackage{subfigure}
\usepackage{amsmath}
\usepackage{url}
\usepackage[linesnumbered,ruled,vlined,boxed]{algorithm2e}
\usepackage{multirow}
\usepackage{placeins}

\usepackage{tikz}
\usetikzlibrary{positioning,fit}
\usetikzlibrary{matrix}
\usetikzlibrary{calc}
\usepackage{ifthen}
\usepackage{pgf}
\usepackage{pgffor}
\usepgfmodule{shapes}
\usepgfmodule{plot}
\usetikzlibrary{decorations}
\usetikzlibrary{arrows}
\usetikzlibrary{snakes}
\usepackage{pgfplots}
\usepackage{standalone}

\definecolor{darkgreen}{rgb}{0.0, 0.2, 0.13}
\definecolor{darkolivegreen}{rgb}{0.33, 0.42, 0.18}

\newcommand{\eg}{\textit{e.g.}\ }
\newcommand{\ie}{\textit{i.e.}\ }

\newcommand{\etal}{\textit{et al.}\ }

\author{
Leo Cazenille$^{1}$,
Nicolas Bredeche$^{2}$,
Jos\'{e} Halloy$^{3}$\\

{1} Department of Information Sciences, Ochanomizu University, Tokyo, Japan\\
{2} Sorbonne Universit\'{e}, CNRS, Institut des Syst\`{e}mes Intelligents et de Robotique, ISIR, F-75005 Paris, France\\
{3} Univ Paris Diderot, Sorbonne Paris Cit\'e, LIED, UMR 8236, 75013, Paris, France
}

\title{Automatic Calibration of Artificial Neural Networks for Zebrafish Collective Behaviours using a Quality Diversity Algorithm}

\begin{document}

\flushbottom
\maketitle
\thispagestyle{empty}

\begin{abstract}
During the last two decades, various models have been proposed for fish collective motion. These models are mainly developed to decipher the biological mechanisms of social interaction between animals. They consider very simple homogeneous unbounded environments and it is not clear that they can simulate accurately the collective trajectories. Moreover when the models are more accurate, the question of their scalability to either larger groups or more elaborate environments remains open.
This study deals with learning how to simulate realistic collective motion of collective of zebrafish, using real-world tracking data. The objective is to devise an agent-based model that can be implemented on an artificial robotic fish that can blend into a collective of real fish. We present a novel approach that uses Quality Diversity algorithms, a class of algorithms that emphasise exploration over pure optimisation. In particular, we use CVT-MAP-Elites~\cite{vassiliades2018using}, a variant of the state-of-the-art MAP-Elites algorithm~\cite{mouret2015illuminating} for high dimensional search space. 
Results show that Quality Diversity algorithms not only outperform classic evolutionary reinforcement learning methods at the macroscopic level (i.e. group behaviour), but are also able to generate more realistic biomimetic behaviours at the microscopic level (i.e. individual behaviour).
\end{abstract}


\section{Introduction}

Many models have been proposed for fish collective behaviours and motion~\cite{lopez2012behavioural,sumpter2012modelling,deutsch2012collective}. At an early stage, they were developed to model realistic collective motion in computer simulation~\cite{Reynolds1987}. Nowadays, most of the models are developed to decipher the interaction rules of the animals and not to replicate their behaviour in autonomous agents be them robots or simulations. It is not clear that they can be used to produce a realistic description of fish collective interactions with collective trajectories~\cite{herbert2015turing} similar to the observations. Moreover, most of the models consider an unbounded homogeneous space that could be the case in pelagic conditions but not in bounded and in-homogeneous environments. Only a few models consider the walls of the tanks that have a important effect on the fish~\cite{jeanson2003model,collignon2016stochastic,calovi2018disentangling}. In the robotic context, developing bio-mimetic and realistic fish behavioural models that can be implemented in robots are difficult to develop~\cite{cazenille2017acceptation,cazenille2017automated}. These issues are related: (i) how can we develop models producing good descriptions of fish collective behaviours and (2) that, when used as controllers, allow fully autonomous agents (robots, simulations) to cope with bounded inhomogeneous environments and social interactions?

For this type of question, currently two kind of modelling methods are pursued to simply take into account the tank walls and the social context. The first one is equation-based. Equations for the motion of the individuals are developed and calibrated on experimental data \cite{jeanson2003model,calovi2018disentangling}. It has been shown that they give excellent results for groups of two fish (\textit{Hemmigramus blerei}) in a circular bounded environment \cite{calovi2018disentangling}. It remains to demonstrate that such method is scalable for groups made of more than two individuals and more elaborate set-ups.
The second kind of modelling technique is agent based. For example, we have developed agent based models that take into account bounded in-homogeneous environment and the social context of the fish \cite{collignon2016stochastic,cazenille2017acceptation}. However, agent based models become rapidly complicated as the number of variables and parameters increases. The scalability of this modelling technique remains also an issue.

Here we explore how to develop scalable effective models to generate robot controllers producing realistic collective behaviours. We do not look for understanding specific collective behaviour mechanisms. In recent works, we explored the use of artificial neural network models (multilayer perceptrons) to generate realistic collective motion and trajectories of a group of five zebrafish in a bounded environment~\cite{cazenille2018nn0,cazenille2018nn1}. We compared supervised learning and reinforcement learning techniques to optimise the behaviour of artificial Zebrafish, so that they would match the trajectories obtained from real-world experimental data. In this setup, learning a behavioural model is challenging because of the continuous state and action spaces as well as the lack of a world model. We showed that evolutionary reinforcement learning, \ie a direct policy search method~\cite{sutton2018reinforcement,whiteson2012evolutionary}, can be used to obtain relevant fish trajectories with respect to individual and collective dynamics, and outperforms results obtained by supervised learning. We also showed that while multi-objective evolutionary optimisation using NSGA-III~\cite{yuan2014improved} could provide different results over single objective optimisation using CMA-ES~\cite{auger2005restart}, the overall quality of trajectories generated is limited by the multiple aspects of behavioural dynamics to be captured simultaneously: wall-following, aggregation, individual trajectories and group dynamics. As a result, we showed that while the \textit{global} biomimetic score (\textit{i.e.} the aggregation of all criteria) is improved with these methods, there is no guarantee that \textit{all} behavioural features will be optimised. In other words, generated trajectories may display unrealistic behaviours, such as low alignment between individuals or erratic wall-following behaviours, while matching real world data in term of inter-individual distances.

In order to improve the quality of biomimetic behavioural strategies, we propose to favour exploration over pure optimisation by using Quality-Diversity (QD) algorithms~\cite{pugh2016quality,cully2018quality}. These algorithms are particularly successful in evolutionary robotics problems~\cite{mouret2015illuminating,cully2015robots,duarte2018evolution}, either by improving diversity to overcome deceptive search spaces~\cite{lehman2013effective}, or by generating a large repertoire of solutions instead of just one single solution~\cite{mouret2015illuminating}. In the current setup (Fig.~\ref{fig:workflow}), we enforce diversity to guide the search by exploring trade-offs between overall quality, which results from aggregating different criteria, and unique realistic behavioural traits, which focus on specific behavioural features, in this case: (1) inter-individual distances
between agents, (2) polarisation of the agents in the group, (3) distribution of agent linear speed and (4) probability of presence in the arena.
We use CVT-MAP-Elites~\cite{vassiliades2018using}, a variant of the MAP-Elites algorithm~\cite{mouret2015illuminating} using centroidal Voronoi tessellations to tackle high-dimensional feature spaces. CVT-MAP-Elites makes it possible to explore a range of both diverse and high-performing solutions by partitioning the search space into geometric regions according to features predefined by the user. It is then possible to find solutions that can be very different from one another.

We show that CVT-MAP-Elites outperforms state-of-the-art evolutionary optimisation methods (CMA-ES and NSGA-III) for revealing biomimetic behavioural strategies in a fish collective. Even more interestingly, we show that trajectories generated by individuals obtained with CVT-MAP-Elites are also more realistic (when compared to actual data from the fish) at the \textit{microscopic} scale, with realistic behaviours at the level of the individuals. Quality Diversity algorithms offer a promising alternative to classical evolutionary optimisation and reinforcement learning algorithms with respect to learning biomimetic controller for artificial fish.

\section{Methods}

\subsection*{Experimental set-up} \label{sec:setup}
We apply the same experimental method, fish handling and set-up as in~\cite{cazenille2017acceptation,seguret2017loose,cazenille2018nn0,cazenille2018nn1}. During experiments, fish are placed in an immersed square white plexiglass arena of $1000\times1000\times100$~mm. An overhead camera records a video of the experiment at 15 FPS with a $500\times500$px resolution. It is them analysed to track the fish positions. Experiments were carried out with 10 groups of 5 adult (6-12 months old) wild-type AB zebrafish (\textit{Danio rerio}) in ten 30-minutes trials as in~\cite{cazenille2017acceptation,seguret2017loose}. 
Experiments conduced in this study were performed under the authorisation of the Buffon Ethical Committee (registered to the French National Ethical Committee for Animal Experiments \#40) after submission to the French state ethical board for animal experiments.

\begin{figure*}[h]
\begin{center}
\includegraphics[width=0.70\textwidth]{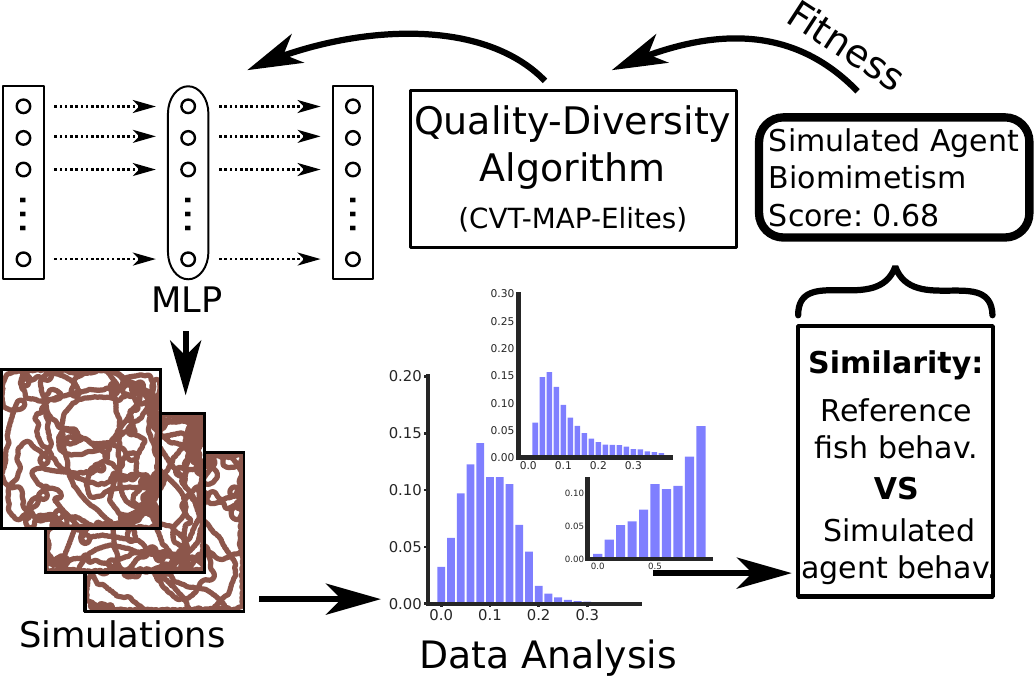}
\caption{Description of the presented methodology to calibrate artificial neural networks to generate fish trajectories. We apply CVT-MAP-Elites~\cite{vassiliades2018using}, a quality-diversity algorithm~\cite{pugh2016quality}, to optimise the weights of a Multilayer Perceptron (MLP, 1 hidden layer, 10 neurons) that drive 5 fish-like agents in simulations. Simulated agents trajectories are compared to experimental fish trajectories. The fitness function corresponds to the biomimetism score of simulated agent groups. CVT-MAP-Elites is compared to CMA-ES~\cite{auger2005restart} as in~\cite{cazenille2018nn0,cazenille2018nn1}. }
\label{fig:workflow}
\end{center}
\end{figure*}

\begin{table*}[h]
\centering
\scriptsize

\begin{tikzpicture}
\node[inner sep=0] (inputs) at (0,0) {
\begin{tabular}{@{} p{3.0cm} p{1.1cm} p{7.5cm} @{}}
\hline
Name & \#Param. & Description \\
\hline
Linear speed & 1 & Instant linear speed of the FA at the prev. time-step\\
Angular speed & 1 & Instant angular speed of the FA at the prev. time-step\\
Distance towards agents & 4 & Linear dist. from the FA towards each other agent\\
Angle towards agents & 4 & Angular dist. from the FA towards each other agent\\
Alignment (angle) & 4 & Angular dist. between the FA heading and other agent heading\\
Alignment (linear speed) & 4 & Difference of linear speed between the FA and other agent linear speed\\
Distance to nearest wall & 1 & Linear dist. from the FA towards the nearest wall\\
Angle towards nearest wall & 1 & Angular dist. from the FA towards the nearest wall\\
\hline
\end{tabular}
};

\node[inner sep=0,below=10mm of inputs] (outputs) {
\begin{tabular}{@{} p{3.0cm} p{1.1cm} p{7.5cm} @{}}
\hline
Name & \#Param. & Description \\
\hline
Delta linear speed & 1 & Change of inst. linear speed of the FA from the prev. time-step\\
Delta angular speed & 1 & Change of inst. angular speed of the FA from the prev. time-step\\
\hline
\end{tabular}
};

\sffamily%\sansmath
\fontsize{12}{25.0}\selectfont
\node[color=black,above left=0mm and 7mm of inputs,shift={(2.00,0.00)}] {\textbf{Inputs}};
\node[color=black,above left=0mm and 7mm of outputs,shift={(2.30,0.00)}] {\textbf{Outputs}};
\end{tikzpicture}

\caption{List of the $20$ parameters used in inputs and of the $2$ parameters used as outputs of the neural network models of agent behaviour. Here, \text{FA} refers to the focal agent.}
\label{tab:inputsOutputs}
\end{table*}

\subsection*{Artificial neural network model} \label{sec:model}
Artificial neural networks (ANN) are universal function approximators able to model phenomena with \textit{a priori} information. They were used in previous studies~\cite{cazenille2018nn0,cazenille2018nn1,iizuka2018learning} to model fish collective behaviour and generate biomimetic trajectories of fish in groups. However this problem is challenging, and it is still possible to improve upon the biomimetism of resulting trajectories. Our methodology builds on Cazenille~\etal~\cite{cazenille2018nn1} and calibrates Multilayer Perceptron (MLP)~\cite{bishop2006prml} artificial neural networks to drive simulated fish-like agents in groups of 5 individuals. All simulations involve 5 simulated agents driven by the optimised MLP (see workflow on Fig~\ref{fig:workflow}).

MLP are a class of feedforward artificial neural networks. They can be employed in a wide variety of modelling and control tasks~\cite{norgaard2000neural}. As in~\cite{cazenille2018nn0,cazenille2018nn1}, our approach uses MLP with one hidden layer of $10$ neurons with a hyperbolic tangent activation function. We use this simple and limited ANN as a baseline for bench-marking the various optimisation algorithms.

Table~\ref{tab:inputsOutputs} lists the parameters used as inputs and outputs of the MLP controllers for each simulated focal agent. The $20$ inputs parameters are often used in multi-agent models of animal collective behaviour~\cite{deutsch2012collective,sumpter2012modelling}, and can arguably be considered to be sufficient to model fish groups trajectories. As we consider fish trajectories observed in a bounded environment, we also take into account the presence of walls, which is often ignored in models of fish behaviour, and only found in a small number of recent studies~\cite{collignon2016stochastic,calovi2018disentangling,cazenille2018nn0,cazenille2017acceptation,cazenille2018nn1}.

\subsection*{Data analysis} \label{sec:dataAnalysis}
As in~\cite{cazenille2018nn0,cazenille2018nn1}, we analyse the tracked positions of agents in each trial $e$ (experiments or simulations) and compute several behavioural metrics: (i) the distribution of \textit{inter-individual distances} between agents ($D_e$); (ii) the distributions of \textit{instant linear speeds} ($L_e$); (iii) the distribution of \textit{polarisation} of the agents in the group ($P_e$); (iv) the \textit{probability of presence of agents in the arena} ($E_e$). The polarisation of an agent group assesses the extent to which fish are aligned. It corresponds to the absolute value of the mean agent heading: $P = \frac{1}{N} \bigl\lvert  \sum^{N}_{i=1} u_i \bigr\rvert$ where $u_i$ is the unit direction of agent $i$ and $N=5$ is the number of agents~\cite{vicsek1995novel}.
Recent studies introduced more complex metrics to assess fish behaviour, like 2D features maps of neighbours compared to a focal fish used in~\cite{jiang2017identifying,heras2018deep}. Our approach here aims to provide a simple methodological baseline, so we only take into account simple and established behavioural metrics like polarisation and inter-individual distances. While more complex metrics based on 2D features maps could describe more accurately fish collective dynamics, they may also require quantities with higher dimensionality than simple metrics, which may make their synthesis into behavioural scores more difficult.

We quantify the realism of the simulated fish-like agents groups by computing a \textbf{biomimetism score} of their behaviour, as in~\cite{cazenille2017acceptation,cazenille2018nn0,cazenille2018nn1}. It measures the similarity between behaviours exhibited by the simulated fish group and those exhibited by the experimental fish averaged across all 10 experimental trials (\textbf{Control} case $e_c$). This score ranges from $0.0$ to $1.0$ and is defined as the geometric mean of the other behavioural scores:
\begin{equation}
S(e, e_c) = \sqrt[4]{I(L_{e}, L_{e_c}) I(D_{e}, D_{e_c}) I(P_{e}, P_{e_c}) I(E_{e}, E_{e_c})}
\end{equation}
The function $I(X, Y)$ is defined as such: $I(X, Y) = 1 - H(X, Y)$.
The $H(X, Y)$ function is the Hellinger distance between two histograms~\cite{deza2006dictionary}. It is defined as: $H(X, Y) = \frac{1}{\sqrt{2}} \sqrt{ \sum_{i=1}^{d} (\sqrt{X_i} - \sqrt{Y_i}  )^2 }$ where $X_i$ and $Y_i$ are the bin frequencies.
As opposed to~\cite{cazenille2018nn1}, we do not take into account the distribution of angular speeds in the computation of the fitness. Indeed, the distributions of angular speeds of evolved individuals was always similar to the ones from random individuals. Thus, we removed this behavioural metrics from the features taken into account to reduce the dimensionality of the feature space.

\subsection*{Optimisation and illumination} \label{sec:optim}

We calibrate the weights of the MLP models driving agent behaviour to approximate as close as possible the trajectories and behaviours of groups of 5 fish-like agents, as in~\cite{cazenille2018nn0,cazenille2018nn1,cazenille2017automated}. Simulations have a duration of 30 minutes ($15$ time-steps per seconds, \ie $27000$ steps per simulation).

In previous studies~\cite{cazenille2018nn0,cazenille2018nn1}, we optimised these MLP controllers using evolutionary algorithms: CMA-ES~\cite{auger2005restart} and NSGA-III~\cite{yuan2014improved}.

Here, we use the CVT-MAP-Elites~\cite{vassiliades2018using} QD algorithm, a variant of the popular MAP-Elites~\cite{mouret2015illuminating} algorithm, to search for interesting MLP controllers matching experimental fish trajectories across a user-provided space of features. The family of Map-Elites algorithms is based on the idea of exploring a clustered search space, retaining the best candidate solutions for each cluster. Clusters correspond to specific range of values for pre-defined features and each candidate solution is stored in a cell of a so-called map, which corresponds to its cluster. The seminal MAP-Elites algorithm uses a pre-defined clustering of the feature space, with the number of clusters (or "bins") quickly exploding as the number of feature dimensions considered grows. In order to tackle high-dimensional feature space, the CVT-MAP-Elite algorithm defines clusters as centroids of Voronoi tesselation, where centroids can be automatically positioned during exploration.

In our case, these features correspond to the four behavioural metrics $L_{e}$, $D_{e}$, $P_{e}$, $E_{e}$ presented earlier.
CVT-MAP-Elites is capable of handling high dimensional feature spaces (like our case) by using centroidal Voronoi tessellations to reduce the dimensionality of the feature space. Here, the CVT-MAP-Elites case only consider $32$ bins of elites, which is far lower as what would be used with MAP-Elites in a reasonable configuration (\eg with 32 bins per features, it would correspond to a grid with $32\times32\times32\times32 = 33554432$ bins of elites).
We selected empirically $32$ bins of elites in the CVT-MAP-Elites methods because it produced the best-performing results among tested numbers of bins.

We compare the generated trajectories using CVT-MAP-Elites with previous results from~\cite{cazenille2018nn0,cazenille2018nn1} where MLP controllers were optimised by the CMA-ES~\cite{auger2005restart}. CMA-ES is a popular mono-objective global optimiser capable of handling problems with noisy, ill-defined fitness function.

In all cases, the algorithms aim to maximise the biomimetism score ($S_{e_o, e_c}$) of MLP-driven agents in simulations ($e_o$) compared to experimental fish groups ($e_c$). Both cases are tested in 10 different trials with the same budget of objective function evaluation (one simulation corresponds to one function evaluation): 60000 evaluations. The CVT-MAP-Elites case involves 6000 evaluations in the initial batch, and 450 batches of 120 individuals. The CMA-ES case involves 500 generations of 120 individuals.

We use a CVT-MAP-Elites implementation from the QDpy (Quality Diversity in Python) framework~\cite{qdpy}.
The CMA-ES implementation is based on the DEAP library~\cite{fortin2012deap}.

%\afterpage{
\begin{figure*}[h]
\begin{center}
\includegraphics[width=0.60\textwidth]{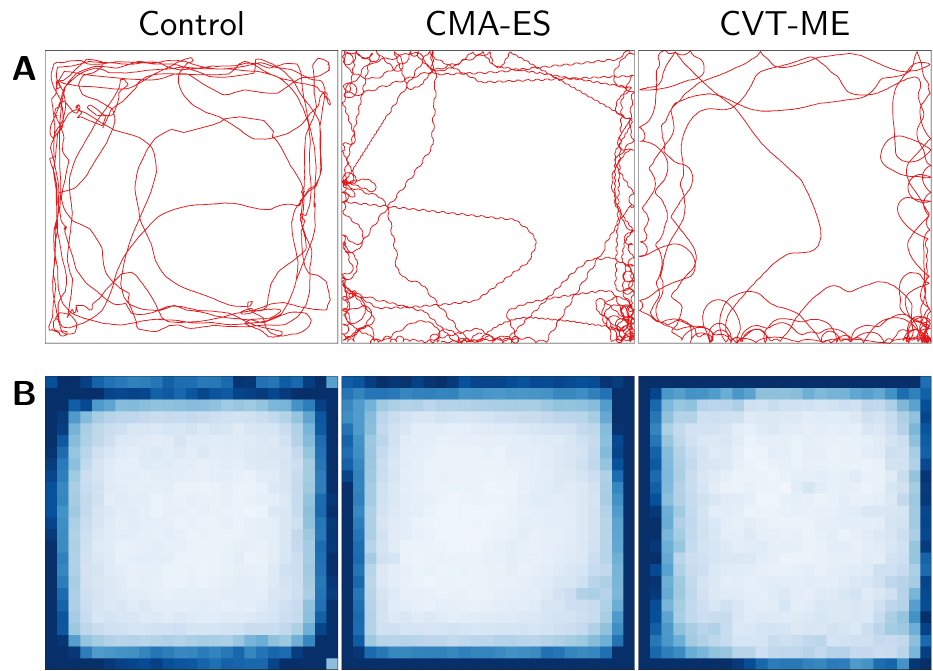}
\caption{Agent trajectories in the square ($1m$) experimental arena after 30-minute trials, for all considered cases: \textbf{Control} reference experimental fish data obtained as in~\cite{collignon2016stochastic,seguret2017loose}, \textbf{CVT-MAP-Elites} and \textbf{CMA-ES} corresponding to simulated MLP-driven agents.
\textbf{A} Examples of an individual trajectory of one agent among the 5 making the group (fish or simulated agent) during 1 minute out of a 30-minute trial. \textbf{B} Presence probability density of agents in the arena.}
\label{fig:plotsTraces}
\end{center}
\end{figure*}
\begin{figure*}[h]
\centering
\includegraphics[width=0.99\textwidth]{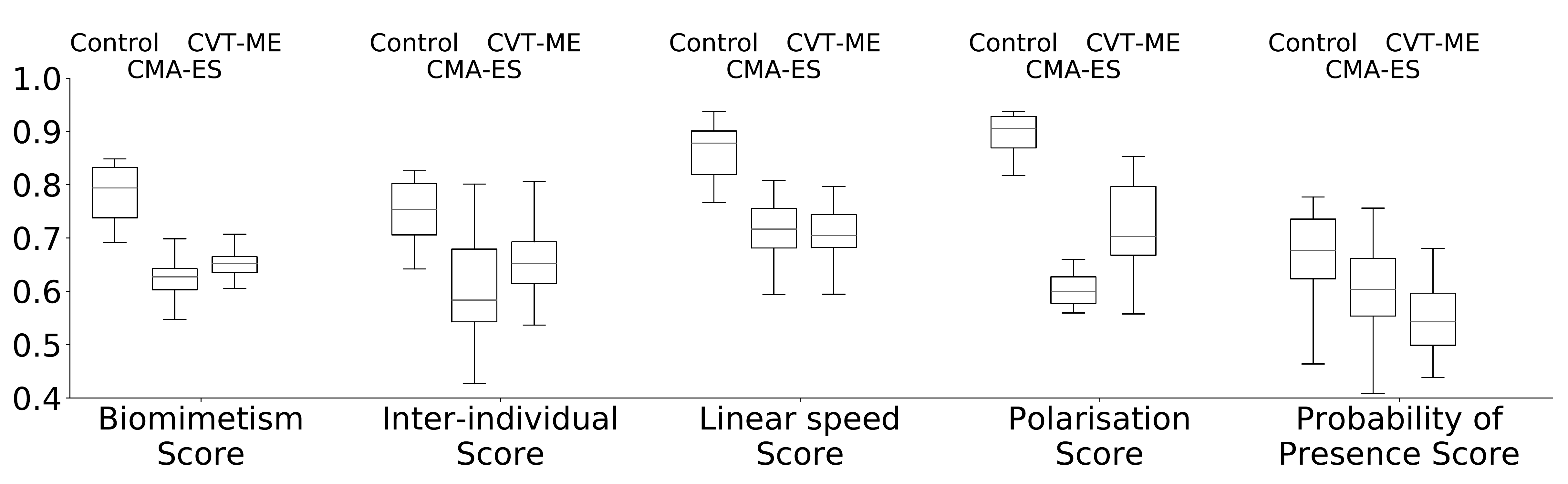}
\caption{Similarity scores between the trajectories of the experimental fish groups (Control) and those of the best-performing simulated individuals optimised by CVT-MAP-Elites or CMA-ES. All cases are tested across 10 different trials (experiments or simulations). Four behavioural features are considered to quantify the realism of exhibited behaviours. \textbf{Inter-individual distances} measures the similarity in distribution of inter-individual distances between all agents and corresponds to the capabilities of the agents to aggregate. \textbf{Linear speed distribution} measures to the distributions of linear speeds of the agents. \textbf{Polarisation} measures how aligned the agents are in the group. \textbf{Probability of presence} corresponds to the density of agent presence in each part of the arena (cf Fig.~\ref{fig:plotsTraces}B). The \textbf{Biomimetic score} is computed as the geometric mean of the other scores. }
\label{fig:scores}
\end{figure*}

\begin{figure*}[h]
\begin{center}
\includegraphics[width=0.70\textwidth]{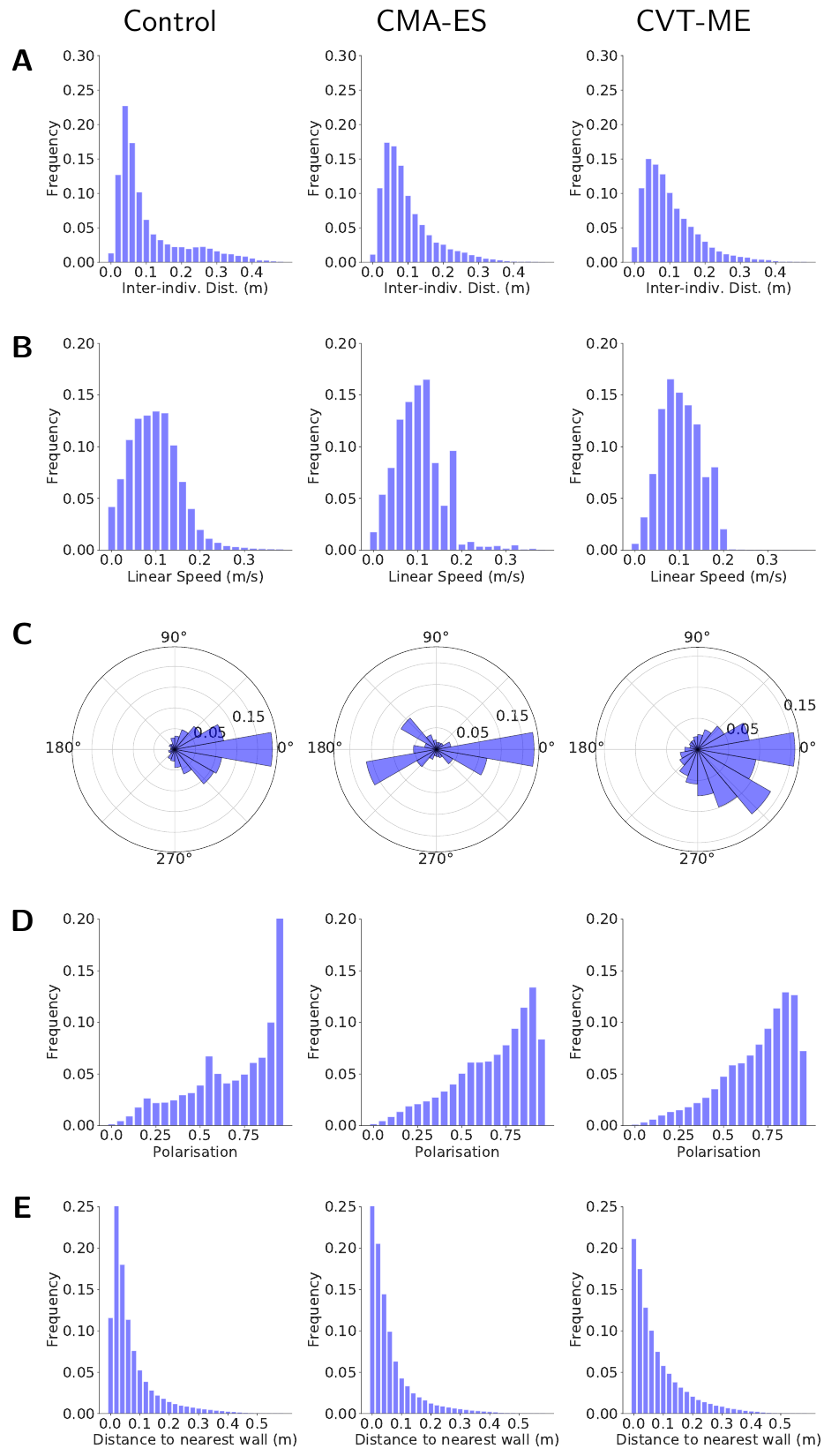}
\caption{Behavioural comparison between ten 30-minute trials of experimental fish in groups of 5 and the 5-sized simulated fish groups for both tested cases. The following behavioural features are examined: inter-individual distances (\textbf{A}), linear (\textbf{B}) and angular (\textbf{C}) speeds distributions, polarisation (\textbf{D}), and distances to nearest wall (\textbf{E}). Note that distributions of angular speeds and distances to nearest wall are informative and not used in the calibration process.}
\label{fig:plotsHists}
\end{center}
\end{figure*}
%}

\section{Results} \label{sec:results}
We analyse the behaviour of the simulated agent groups for the CVT-MAP-Elites and CMA-ES cases and compare them with the behaviour of experimental fish groups (Control case). In both cases, the agents are driven by MLP controllers, calibrated either by CVT-MAP-Elites or with CMA-ES to match as close as possible the behaviour of experimental fish across the behavioural metrics presented above. Each case is repeated in 10 trials and the following statistics only consider the best-evolved MLP controllers.

Figure~\ref{fig:plotsTraces}A provides examples of agents trajectories. In the control case, fish tend to follow walls but retain a capability to go to the center of arena. This is also observed in trajectories from both MLP-driven cases. 
However, they also incorporate patterns not found in actual fish trajectories. Small circular loops can appear in both cases. A small periodic "shaking" is present in the trajectories of the CMA-ES case. Conversely, the trajectories of the CVT-MAP-Elites appear smoother and match more closely those of the experimental fish. This suggests that CVT-MAP-Elites is more realistic at the microscopic level of agent trajectories.
Figure~\ref{fig:plotsTraces}B presents the mean probability of presence of all agents in the arena for all cases.

We assess the realism of the two tested cases by computing the behavioural metrics presented in Sec.~\ref{sec:dataAnalysis}. These metrics serve as a base to compute similarity scores between the tested cases and experimental fish behaviour (Fig.~\ref{fig:scores}).  Both simulated cases display lower similarity scores than the experimental fish groups.
Based on a comparison of the best solutions found by both algorithms, CVT-MAP-Elites outperforms CMA-ES with statistical significance (p-value=$0.0227$ using the Mann-Whitney U-test). The best solution found by CVT-MAP-Elites also dominates all solutions found with CMAE-ES (best fitness: $0.724$ with CVT-MAP-Elites \textit{vs.} $0.704$ with CMA-ES).

However, the controllers optimised by the two methods prioritise different features. The CVT-MAP-Elites case shows higher scores on inter-individual distances and polarisation than the CMA-ES case. In turn, CMA-ES exhibits higher probability of presence scores than the CVT-MAP-Elites case. Scores of linear speeds are roughly similar between the two cases. Overall, it means that the controllers optimised by the two methods exhibit different kind of behaviours and way of coping with the trade-offs between fish aggregative and wall-following behaviours. In term of group dynamics, the solutions of the CVT-MAP-Elites case are more cohesive than what is seen in the CMA-ES case, which evolves controllers that are more biased towards wall-following than group aggregation.

Histograms of all behavioural metrics are shown for all cases in Fig.~\ref{fig:plotsHists}, with two complementary metrics: the distribution of angular speeds (Fig.~\ref{fig:plotsHists}, related to polarisation) and distance to nearest wall (Fig.~\ref{fig:plotsHists}, related to probability of presence). They confirm the results from Fig.~\ref{fig:scores}.
The distributions of angular speed (Fig.~\ref{fig:plotsHists}C) of both cases are sub-optimal in term of realism.
Figure~\ref{fig:plotsHists}E displays that simulated agents of both cases tend to exhibit correctly a wall-following behaviour. 

The experimental fish groups of the \textbf{Control} case display a large behavioural variability across all investigated metrics (Fig.~\ref{fig:scores} and~\ref{fig:plotsHists}). Indeed, experiments were conduced with 10 groups of 5 fish (totalling 50 different fish) displaying disparate behaviours and individual preferences. This matches results from previous zebrafish collective behaviours studies~\cite{seguret2017loose,collignon2017collective}. Social (group composition) and environmental contexts impact fish behaviour: fish tend to aggregate in small short-lived sub-groups that follow walls from a distance that vary according to group composition. They also tend to exhibit an uniform degree of alignment within sub-groups.

\section{Discussion and Conclusion} \label{sec:conclusion}

Calibrating artificial neural networks to model the collective behaviour of fish group and generate realistic fish trajectories is a challenging problem because fish behaviours involve several complementary dynamics with trade-offs between group-level dynamics (aggregative tendencies, group alignment), individual-level behaviours (agent linear speed) and response to environmental cues (wall-following behaviour, probability of presence in the arena). It is difficult to balance these conflicting behaviours during the calibration process.

Here, we show that the CVT-MAP-Elites~\cite{vassiliades2018using}, a quality diversity method that emphasises exploration over pure optimisation, calibrates controllers that are more realistic in term of agent groups polarisation and inter-individual distances when compared to previous results using stochastic optimisation methods such as the CMA-ES evolutionary method~\cite{cazenille2018nn0,cazenille2018nn1}. Moreover, QD algorithms also have the advantage of exploring a range of diverse solutions instead of searching for a single local optimum, and could be used to decipher the interrelation between features and behavioural biomimetism in order to draw biological conclusions.

Our approach could still be improved further, either by taking into account more behavioural metrics (tangential and normal accelerations, curvature or tortuosity) or by using more complex artificial neural networks than MLP, such as recurrent neural networks or deep neural networks.

Additionally, our methodology could be adapted to make possible to derive biological conclusions from the calibrated ANN models. ANN can be used as benchmarks to find the necessary information in experimental data to replicate experimental fish behaviour. Recently, Heras \etal~\cite{heras2018aggregation} hinted at the possibility of this approach to decipher the interaction mechanism in large zebrafish groups. It remains to be shown that such ANN models can also produce collective trajectories similar to those observed experimentally. If it is shown to be the case, best-performing agents optimised through such methodology could be used as controllers to drive the behaviour of robots interacting experimentally with fish to study their collective dynamics.

\section*{Acknowledgement}
{\small This work was funded by EU-ICT project 'ASSISIbf', no 601074. 
}

\FloatBarrier
%\clearpage
%\bibliography{biblio}
%\bibliographystyle{unsrt}

\end{document}